\documentclass[twocolumn]{autart}

\newtheorem{theorem}{Theorem}

\newtheorem{proposition}[theorem]{Proposition}

\newtheorem{lemma}[theorem]{Lemma}

\newcommand{\R}{{\mathbb R}}

\newcommand{\Diag}{\mbox{diag}}

\newcommand{\Th}[1]{\( {\it #1}  \)-th}
\newcommand{\Tr}{\mathrm{Tr}}
\newcommand{\GCV}{\mathrm{GCV}}

\RequirePackage{amsmath, amssymb, graphicx,xspace,epstopdf}
\newtheorem{ex}{Example}

\DeclareMathOperator*\argmin{arg\,min}

\usepackage{graphicx}
\usepackage{epsfig} 
\usepackage{mathptmx} 
\usepackage{times} 
\usepackage{amsmath} 
\usepackage{amssymb}  
\usepackage{amsfonts}
\usepackage{euscript}
\usepackage{color}
\allowdisplaybreaks

\begin{document}

\begin{frontmatter}

\thanks[footnoteinfo]{This research has been partially supported by
the MIUR FIRB project RBFR12M3AC-Learning meets time: a new
computational approach to learning in dynamic systems and by the
Progetto di Ateneo CPDA147754/14-New statistical learning approach
for multi-agents adaptive estimation and coverage control.
This paper was
not presented at any IFAC meeting. Corresponding author Gianluigi
Pillonetto Ph. +390498277607.}

\title{The Generalized Cross Validation Filter}

\author[second]{Giulio Bottegal} and
\author[first]{Gianluigi Pillonetto}
 
\address[second]{Department of Electrical Engineering, TU Eindhoven, Eindhoven, The Netherlands (e-mail: g.bottegal@tue.nl)}
\address[first]{Department of Information  Engineering, University of Padova, Padova, Italy (e-mail: giapi@dei.unipd.it)}


\maketitle
\begin{abstract}
\emph{Generalized cross validation} (GCV) is one of the most important approaches 
used to estimate parameters in the context of inverse problems and regularization techniques.
A notable example is the determination of the smoothness parameter
in splines. When the data are generated by a state space model,
like in the spline case, efficient algorithms are available to 
evaluate the GCV score with complexity that scales linearly in the data set size. 
However, these methods are not amenable to on-line applications
since they rely on forward and backward recursions. 
Hence, if the objective has been evaluated at time $t-1$ and new data arrive at time
$t$, then $O(t)$ operations are needed to update the GCV score.
In this paper we instead show that 
the update cost is $O(1)$, thus paving the way to the on-line use of GCV.
This result is obtained by deriving the novel \emph{GCV filter} 
which extends the classical Kalman filter equations to efficiently propagate
the GCV score over time. We also illustrate applications of the new filter in the context of state estimation
and on-line regularized linear system identification.
\end{abstract}

\end{frontmatter}

\section{Introduction}\label{Sec1}

Linear state space models assume the form
\begin{subequations}
\begin{align*}
x_{k+1} &= A_k x_k  + \omega_k \\ \label{Meas}
y_k  &= C_k x_k + e_k 
\end{align*}
\end{subequations}
where $x_k$ is the state at instant $k$, $y_k$ is the output, while
$\omega_k$ and $e_k$ are random noises. 
The matrices $A_k$ and $C_k$ regulate the state transition and the observation model at instant $k$. 
This kind of models plays a central role in the analysis and design of discrete-time systems \cite{Kalman60}. 
Applications abound and include tracking, navigation and biomedicine.\\

In  \emph{on-line state estimation}, the problem is the reconstruction
of the values of $x_k$ from measurements of $y_k$ collected over time. 
When the matrices $A_k$ and $C_k$ and
the noises covariances are known, the optimal linear 
estimates are efficiently returned by 
the classical Kalman filter \cite{Anderson:1979}. However, in many circumstances
there can be unknown model parameters 
that also need to be inferred from data in an on-line manner, 
e.g. variance components or entries of the transition/observation matrices. 
One can interpret such parameters as additional states.
Then, the extended Kalman filter \cite{Jaz70} or
more sophysticated stochastic techniques, such as
particle filters and Markov chain Monte Carlo \cite{Gilks,Ninness2010,ParticleMCMC,Frigola2013a}, can be used to track the filtered posterior.
Another technique consists of propagating the marginal likelihood of the unknown
parameters via a bank of filters \cite[Ch. 10]{Anderson:1979}. In this paper, we will show that another viable alternative 
is the use of an approach known in the literature as generalized cross validation (GCV) \cite{Golub79}.\\

In the literature of statistics and inverse problems, GCV is widely 
used in off-line contexts to estimate unknown parameters
entering regularized estimators \cite{Bertero1,Tarantola2005,Wahba1990}.  
This approach was initially used to tune the smoothness parameter
in ridge regression and smoothing splines \cite{Hoerl1970,Golub79,Rice1986}. GCV is now also popular 
in machine learning, used to improve the generalization capability
of regularized kernel-based approaches \cite{Scholkopf01b,PoggioMIT}, such as regularization networks, which contain spline regression as special case \cite{Poggio90,Girosi95}.\\
 {To introduce GCV in our state space context, we first recall that smoothing splines are closely linked to state space models of $m$-fold integrated Wiener processes \cite{pillonetto2009fast}; then it appears natural to extend GCV to general state space models.} To this end,
assume that measurements $y_k$ have been collected up to instant $t$ and
stacked in the vector $Y_t$. Denote with $\hat{Y_t}$ the vector containing the optimal linear output estimate\footnote{The 
components of $\hat{Y_t}$ are thus given by $C\hat{x}_{k | t}$, where the smoothed state $\hat{x}_{k | t}$ 
can be obtained for any $t$ with $O(t)$ operations by a fixed-interval Kalman smoothing filter \cite{RTS,Ljung:1976}.} and
use $H_t$ to denote 
the \emph{influence matrix} satisfying  
$$
\hat{Y_t} = H_t Y_t.
$$
Then, the parameter estimates achieved by GCV minimize 
\begin{equation}\label{GCVintro}
\GCV_t = \frac{S_t}{t(1-\delta_t/t)^2},
\end{equation}
where $S_t$ is the sum of squared residuals, i.e. 
$$S_t=\|\hat{Y_t} - Y_t\|^2,$$
and  $\delta_t$ are the \emph{degrees of freedom} given by the trace of $H_t$, i.e. 
$$\delta_t=\Tr(H_t).$$
In the objective (\ref{GCVintro}), the term $S_t$ accounts for the goodness of fit while 
$\delta_t$ assumes values on $[0,t]$ and measures model complexity. In fact,  
in nonparametric regularized estimation, the degrees of freedom $\delta_t$
can be seen as the counterpart of the number of parameters entering a parametric model \cite{MacKay,Hastie01,MLAuto2015}.\\
GCV is supported by important asymptotic results. 
Also, for finite data set size it turns often out a good approximation
of the output mean squared error \cite{Craven79}.
It is worth stressing that such properties have been derived 
without postulating the correctness of the prior models describing the output data \cite{Wahba83,Wahba85}.
In control, this means that GCV can compensate for possible
modeling mismatch affecting the state space description.

Despite these nice features, the use of
GCV 
within the control community appears limited, in particular in on-line contexts. 
One important reason is the following one.
For state space models, there exist efficient algorithms
which, for a given parameter vector, return its
GCV score with $O(t)$ operations \cite{Kohn89,ansley1987efficient},
see also  \cite{Hutchinson85,Silverman85,DeNic2000} for procedures
dedicated to smoothing splines.
But all of these techniques are not suited to on-line computations
since they involve forward and backward recursions. 
Hence, if $\GCV_{t-1}$ 
is available and new data arrive at time
$t$, other $O(t)$ operations are needed to achieve $\GCV_t$.
In this paper, we will instead show that 
the update cost is $O(1)$, thus paving the way to a more pervasive 
on-line  use of 
GCV. 
This result is obtained by deriving the novel \emph{GCV filter}
which consists of an extension of the classical Kalman equations. 
Thanks to it, one can run a bank of filters (possibly in parallel)
to efficiently propagate GCV over a grid of parameter values.  {This makes the proposed GCV filter particularly suitable for applications where a measurement model admits a state space description with dynamics depending on few parameters, see e.g. the next section for an application in 
numerical differentiation. In this framework, an implementation of the GCV filter via a bank of parallel filters turns out computationally attractive.}

The paper is organized as follows. In Section \ref{Sec2}, first
some additional notation is introduced. Then, 
the GCV filter is presented. 
Its asymptotic properties are then discussed in Section \ref{Sec3}.  
In Section \ref{Sec4} we illustrate some applications, including also smoothing splines and  
on-line regularized linear system identification
with the stable spline kernel used as stochastic model for the impulse response \cite{SS2010,SurveyKBsysid}.
Conclusions end the paper while the correctness of the GCV filter is shown
in Appendix.

\section{The GCV filter}\label{Sec2}

\subsection{State space model}

First, we provide full details about our 
measurements model. We use $x \sim (a,b)$
to denote a random vector $x$ with mean $a$ and
covariance matrix $b$. Then, our state space model is 
defined by
\begin{subequations}\label{StateMod}
\begin{align}\label{State}
x_{k+1} &= A_k x_k  + \omega_k \\ \label{Meas}
y_k  &= C_k x_k + e_k, \ \ k=1,2,\ldots\\
x_1 & \sim (\mu,P_0) \\
\omega_k & \sim (0,Q_k) \\
e_k & \sim (0,\gamma)  
\end{align}
\end{subequations} 
where the initial condition $x_1$ 
and all the nosies $\{\omega_k,e_k\}_{k=1,2,\ldots}$ are mutually uncorrelated.  {We do not specify any particular distribution for these variables, since the GCV score does not depend on the particular noise distribution\footnote{Of course, GCV may result not effective if the noises are highly non-Gaussian. Different approaches, like particle filters, should instead be used if linear estimators perform poorly due e.g. to multimodal noise distributions.}. If $x_1,\,\omega_k,\,e_k$ are Gaussian, then the Kalman filter provides the optimal state estimate in the mean-square sense. In the other cases, the Kalman filter corresponds to the best linear state estimator \cite{Anderson:1979}.}
In addition, just to simplify notation the measurements $y_k$
are assumed scalar, so that $\gamma$ represents the noise variance.\\
We assume that some of the parameters in \eqref{StateMod} may be unknown, or could enter $A_k,B_k,Q_k$ and $P_0$; however, we 
do not stress this possible dependence to make the formulas more readable. 
The matrix $P_0$ is assumed to be independent of $\gamma$.  
Such parameter is typically unknown, being connected to the ratio between the measurement noise variance and the variance of the driving noise.
It corresponds to the regularization parameter in the smoothing-splines context described in the example below.

 {\begin{ex}[Smoothing splines \cite{PillSacc}] \label{SplineEx}
Function estimation and numerical differentiation are often required in various applications. 
These include also input reconstruction in nonlinear dynamic systems as described e.g. in \cite{PillSacc}. 
Assume that one is interested in determining the first $m$ derivatives 
of a continuous-time signal measured 
with non-uniform sampling periods $T_k$. Modeling the signal as an \Th{m} fold integrated Wiener process
one obtains the stochastic interpretation of the $m$-th order smoothing splines \cite{Wahba1990}.
In particular, one can use \eqref{StateMod} to represent the signal dynamics as follows
\begin{align*}
A_k  & = \left(\begin{matrix}
1 & 0 & 0 & \ldots & 0 \\
T_k & 1 & 0 &\ldots  & 0  \\
\frac{T_k^2}{2} & T_k  & \ddots & \ddots & \vdots \\
\vdots & \vdots & \ddots & \ddots &  \vdots \\
\frac{T_k^m}{m!} & \frac{T_k^{m-1}}{(m-1)!} &\ldots &T_k & 1
\end{matrix}\right) \quad,\quad
C_k  = \left(\begin{matrix}
0 \\ 0 \\ \vdots \\ 1
\end{matrix}\right)^T, \\
[Q_k]_{ij} & = \frac{T_k^{i+j-1}}{(i-1)!(j-1)!(i+j-1)} \,.
\end{align*}
Such model depends on 
the measurement noise variance $\gamma$, making this application particularly suited for the GCV filter.
\end{ex}}

\subsection{The GCV filter}

The GCV filter equations are now reported.
Below, $\hat{x}_k$ denotes the optimal linear one-step ahead state prediction having
covariance $P_k$.
Its dynamics are regulated by the classical Kalman filter 
via (\ref{Kgain}), (\ref{xpred}) and the Riccati equation (\ref{DRE}).  

{\bf{GCV filter}}

{\emph{Initialization}}
\begin{subequations}
\begin{align}
\hat{x}_{1} &= \mu, \quad \hat{\zeta}_{1} = 0 \\
P_{1} &= P_0, \quad \Sigma_{1} = 0 \\
\delta_{1} &= 1-\gamma(C_1 P_0 C_1^T +\gamma)^{-1} \\
S_{1} &= \gamma^2\frac{(y_{1} - C_1 \mu)^2}{(C_1 P_0 C_1^T +\gamma)^{2}} \\
\GCV_{1} &= \frac{S_{1}}{(1-\delta_{1})^2} 
\end{align}
\end{subequations}
{\emph{Update}}
\begin{subequations}
\begin{align}
\label{Kgain} K_{k} &= A_{k}P_{k}C_{k}^T  (C_{k} P_{k} C_{k}^T +\gamma)^{-1} \\
\label{Ggain}  G_{k} &= \frac{A_{k} \Sigma_{k} A_{k}^T - K_{k}(C_{k} \Sigma_{k} C_{k}^T + 1)}{C_{k} P_{k} C_{k}^T +\gamma}  \\
\label{xpred} \hat{x}_{k+1} &= A_{k} \hat{x}_{k} + K_{k}(y_{k} - C_{k} \hat{x}_{k} )\\
\label{zpred}  \hat{\zeta}_{k+1} &= (A_{k}-K_{k}C_{k}) \hat{\zeta}_{k} + G_{k}(y_{k} - C_{k} \hat{x}_{k} )\\ 
\label{DRE}  P_{k+1} &=(A_{k}-K_{k}C_{k})  P_{k}  (A_{k}-K_{k}C_{k})^T  + \gamma K_{k} K_{k}^T + Q_{k} \\
\label{DRE2}  \Sigma_{k+1} &=(A_{k}-K_{k}C_{k})  \Sigma_{k}  (A_{k}-K_{k}C_{k})^T  + K_{k} K_{k}^T \\
\label{DofRec} \delta_{k+1} &= \delta_{k} + 1-\gamma  \frac{C_{k+1} \Sigma_{k+1} C_{k+1}^T + 1}{C_{k+1} P_{k+1} C_{k+1}^T +\gamma} \\
\label{SsrRec} S_{k+1} &= S_{k} 
                 + \gamma^2  \frac{C_{k+1} \Sigma_{k+1} C_{k+1}^T + 1}{(C_{k+1} P_{k+1} C_{k+1}^T +\gamma)^{2}}(y_{k+1} - C_{k+1} \hat{x}_{k+1})^2 \\ \nonumber
                 &\quad +  2\gamma^2 C_{k+1} \hat{\zeta}_{k+1} \frac{y_{k+1} - C_{k+1} \hat{x}_{k+1}}{C_{k+1} P_{k+1} C_{k+1}^T +\gamma} \\
\label{GcvRec} \GCV_{k+1} &= (k+1)\frac{S_{k+1}}{(k+1-\delta_{k+1})^2}  
\end{align}
\end{subequations}

\begin{figure}
\center
{\includegraphics[scale=0.5]{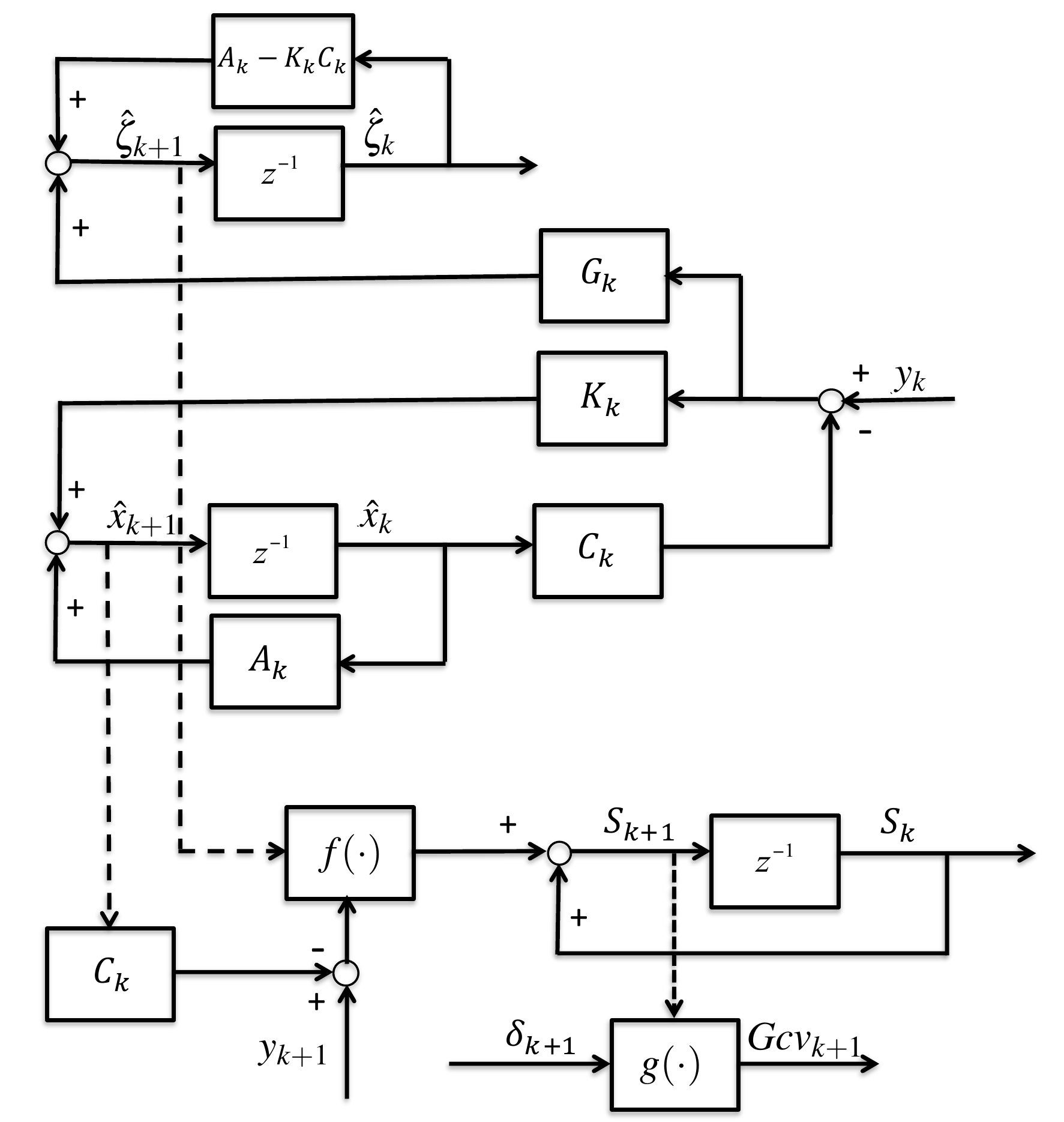}}  
\caption{\emph{GCV filter:} in the bottom the nonlinear blocks $f$ and $g$ are
defined, respectively, by
(\ref{SsrRec}) and (\ref{GcvRec}) while $\delta_{k+1}$ can be recursively computed by  (\ref{DofRec}).}
\label{FigGCV}
\end{figure}

It is apparent that the difference w.r.t the classical Kalman filter is the presence of the additional state 
$\hat{\zeta}_{k}$ of the same dimension of $\hat{x}_{k}$. 
Comparing (\ref{xpred}) and (\ref{zpred}), one can see that $A_{k}$ is replaced by $A_{k}-K_{k} C_{k}$.
In addition, the dynamics of $\hat{\zeta}_{k}$ are still driven by the innovation
$y_{k} - C_{k} \hat{x}_{k}$, but the Kalman gain $K_k$ given by (\ref{Kgain}) is substituted by 
the $G_k$ defined by (\ref{Ggain}).
In turn, such gain depends on $\Sigma_k$ which is propagated over time through a modified version of the Riccati
equation given by (\ref{DRE2}). The GCV filter is graphically depicted in Fig. \ref{FigGCV}.

\section{Asymptotic behavior and the smoothing ratio}\label{Sec3}

\subsection{Asymptotic behavior of the GCV filter}
We first consider the case where the state-space model \eqref{StateMod} is time-invariant, i.e. the matrices $A_k$, $C_k$, and $Q_k$ are constant in $k$. The structure of the equations governing the GCV filter 
permits to easily understand its asymptotic behaviour. 
In particular, exploiting well known properties of the Kalman filter \cite{Anderson:1979},
the following result is obtained (see the Appendix for a proof).

\begin{proposition}\label{GCVregime}
Assume that the system (\ref{StateMod}) is time-invariant, stabilizable and detectable.
Then, for any $P_0$ we have
 \begin{equation*}
\lim_{k \rightarrow \infty} P_k = \bar{P} \ \ \mbox{and} \  \ \lim_{k \rightarrow \infty} \Sigma_k = \bar{\Sigma} 
 \end{equation*}
where $\bar{P}$ and  $\bar{\Sigma}$ are the unique symmetric and semidefinite positive matrices
solving, respectively, the algebraic Riccati equation
\begin{equation}\label{ARE}
\bar{P} = A\bar{P}A^T + Q - A\bar{P}C^T(C \bar{P} C^T+\gamma)^{-1}C\bar{P} A^T
\end{equation}
and the Lyapunov equation
\begin{equation}\label{ARE2}
\bar{\Sigma} =(A-\bar{K}C)  \bar{\Sigma} (A-\bar{K}C)^T  + \bar{K} \bar{K}^T 
\end{equation}
where $\bar{K}=A\bar{P}C^T  (C \bar{P} C^T +\gamma)^{-1}$.
In addition, all the roots of the matrix 
$A-\bar{K}C$ are inside the unit circle so that the (asymptotic) GCV filter is 
asymptotically stable.
\end{proposition}
Properties of the GCV filter can be also characterized in the time-varying case.
In particular, following Section 2 of \cite{And1981}, one can first replace stabilizability and detectability with the assumptions
of uniform stabilizability and detectability. Then, following the same reasonings contained in the proof of Proposition \ref{GCVregime},
Theorem 5.3 in  \cite{And1981} ensures the uniform exponential stability of the GCV filter.

\subsection{Fast regularization parameter tuning and the smoothing ratio}
Proposition \ref{GCVregime} leads also to a new computationally appealing approach to tune the regularization parameter  
$\gamma$ e.g. in smoothing splines. In particular, consider the scenario 
described in \cite{DeNic2000} where an unknown function has to be reconstructed
by spline regression from equally spaced noisy samples. 
When assumptions in Proposition \ref{GCVregime} hold true,   
it is possible to compute off-line the gains
$$
\bar{K}=A\bar{P}C^T  (C \bar{P} C^T +\gamma)^{-1}, \quad  \bar{G}=\frac{A \bar{\Sigma} A^T - \bar{K}(C \bar{\Sigma} C^T + 1)}{C \bar{P} C^T +\gamma}. 
$$
Then, one can exploit the asymptotic (suboptimal) GCV filter,
with the guarantee that the objective values will converge to the exact GCV scores as $k$ increases. 
Moreover, in off-line contexts this approach appears computationally appealing
even when compared to the many GCV-based spline algorithms developed in the last decades  
\cite{Weinert13,Wahba1990,Kohn89,Hutchinson85,Silverman85}.\\
Furthermore, \cite{DeNic2000}
defined the \emph{asymptotic smoothing ratio} as 
$$
\lim_{k \rightarrow \infty} \ \frac{\delta_k}{k},
$$
also providing an interesting closed-form expression for the 
cubic splines case useful to further speed up the tuning of $\gamma$.
For the general case, we notice that 
Proposition \ref{GCVregime} gives also 
a numerical procedure to compute the asymptotic smoothing ratio (for different values of $\gamma$). 
In fact, if (\ref{StateMod}) is stabilizable and detectable, 
combining (\ref{DofRec}) and Proposition \ref{GCVregime} we obtain
$$
\lim_{k \rightarrow \infty} \ \frac{\delta_k}{k}  =  1-\gamma  \frac{C \bar{\Sigma} C^T + 1}{C \bar{P} C^T +\gamma} 
$$
with $\bar{\Sigma}$ and $\bar{P}$ defined, respectively,  in (\ref{ARE}) and (\ref{ARE2}).

\section{Numerical Examples}\label{Sec4}
\subsection{Spline example}\label{Splinesub}

We consider the reconstruction of the function $\exp(\sin 8t)$ taken from \cite{SurveyKBsysid}
from samples collected at 400 instants $t_i$ randomly generated from a uniform distribution
on $[0,1]$. The measurement noise is Gaussian with standard deviation equal to 0.3.
We model $f$ as the two-fold integral of white noise setting
$m=2$ in the time-varying state space model reported in Example \ref{SplineEx}.
This corresponds to reconstructing $f$ using cubic smoothing splines \cite{Wahba1990}.\\ 
We use $Z_t$ to denote the vector containing the 
noiseless outputs (corresponding to the second entries of $\{x_k\}_{k=1}^{t}$). 
We denote the average of $Z_t$ by the scalar quantity $\bar{Z}_t$.
Then, the performance measure is 
the percentage fit 
\begin{equation}\label{Fit}
\mathcal{F}_t = 100\% \left(1-\frac{\|Z_t-\hat{Z}_t\|}{\|Z_t- 1\!\!\!1\bar{Z}_t\|}\right),
\end{equation}
 {where $\hat Z_t$ is the estimate of $Z_t$ obtained through the Kalman smoother \cite{Anderson:1979}, and $1\!\!\!1$ a column vector with all entries equal to 1}. The following two different estimators $\hat{Z}_t$ are tested:
\begin{itemize}
\item \emph{GCV:} this approach 
estimates $\gamma$ exploiting the GCV filter.
More specifically, the GCV score is propagated over a grid containing 
100 values of $\gamma$ logarithmically spaced on the interval $[10^{-2},10^4]$.
Then, at any $t$ the estimate $\hat{Z}_t$ is computed by a Kalman smoothing filter
which exploits the $\gamma_t$ that minimizes $GCV_t$.
\item \emph{Oracle:} the same as GCV except that $\gamma_t$ 
maximizes the fit $\mathcal{F}_t$. Note that this approach
is not implementable in practice since it uses an oracle that knows the noiseless (unavailable) output $Z_t$.
\end{itemize}

\begin{figure*}[htbp]
  \begin{center}
   \begin{tabular}{cccc}
\hspace{.1in}
 { \includegraphics[scale=0.46]{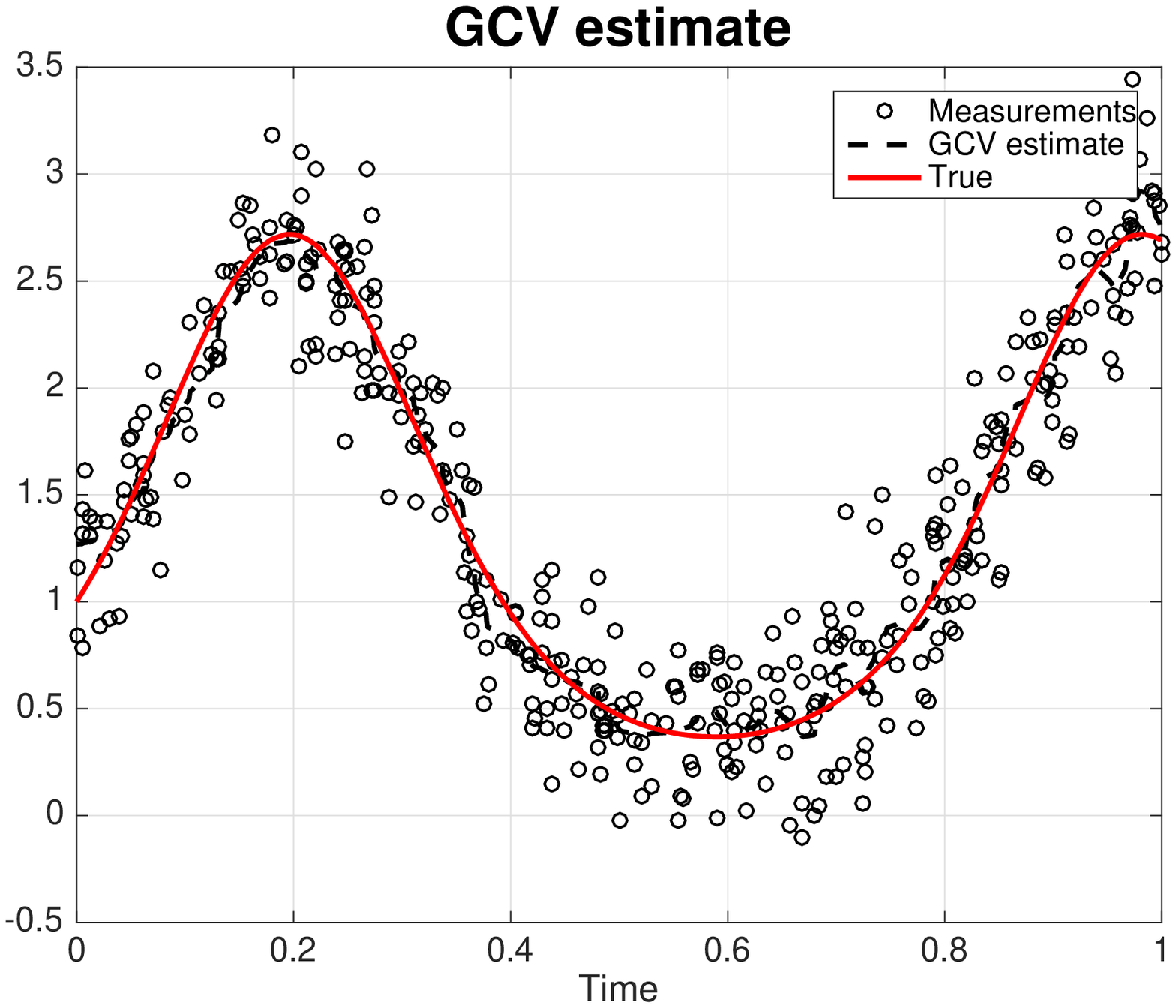}} 
\hspace{.1in}
 { \includegraphics[scale=0.46]{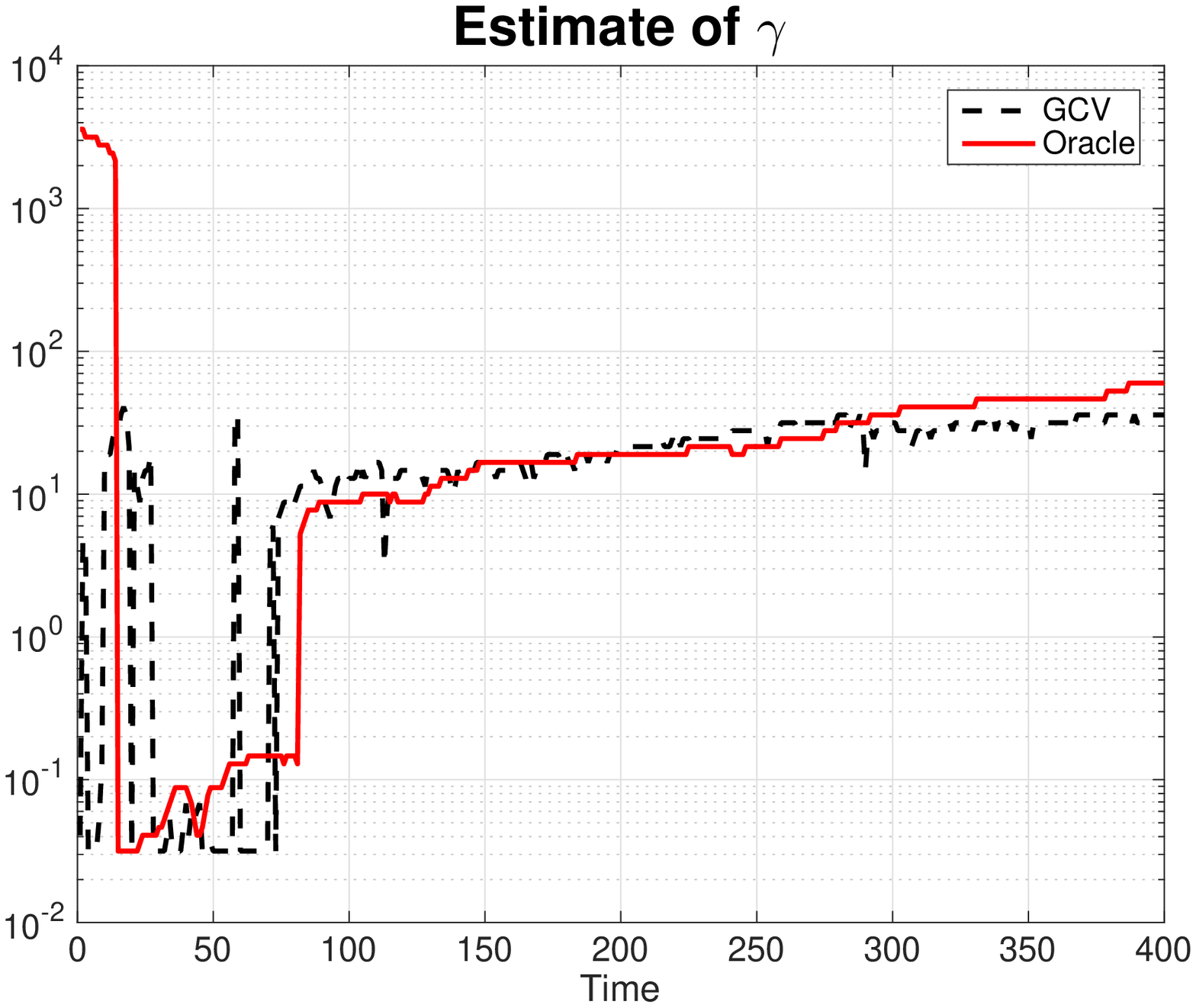}} 
    \end{tabular}
 \caption{{\bf{Cubic spline example - Section \ref{Splinesub}}}. {\it{Left:}} noiseless output (solid line), measurements  ($\circ$) and 
 cubic spline estimate obtained by GCV (dashed line). 
 {\it{Right:}} Estimated regularization parameter $\gamma_t$, as a function of time, obtained by Oracle maximizing the fit 
 $\mathcal{F}_t$ in eq. \ref{Fit} (solid line) and by GCV minimizing the score $GCV_t$ computed by
 eq. \ref{GcvRec}  (dashed line).} 
    \label{FigSpline}
     \end{center}
\end{figure*}

The left panel of Fig. \ref{FigSpline} displays the noiseless output (solid line), the measurements  ($\circ$) 
and the function estimate returned by GCV (dashed line) which turns out close to $f$. 
The right panel also shows that the GCV filter is able to track well and in an on-line manner 
the time-course of $\gamma$ returned by \emph{Oracle}. 

\subsection{GCV capability to compensate for model mismatch}\label{DCmotorsub}

\begin{figure*}[htbp]
  \begin{center}
   \begin{tabular}{cccc}
\hspace{.1in}
 { \includegraphics[scale=0.46]{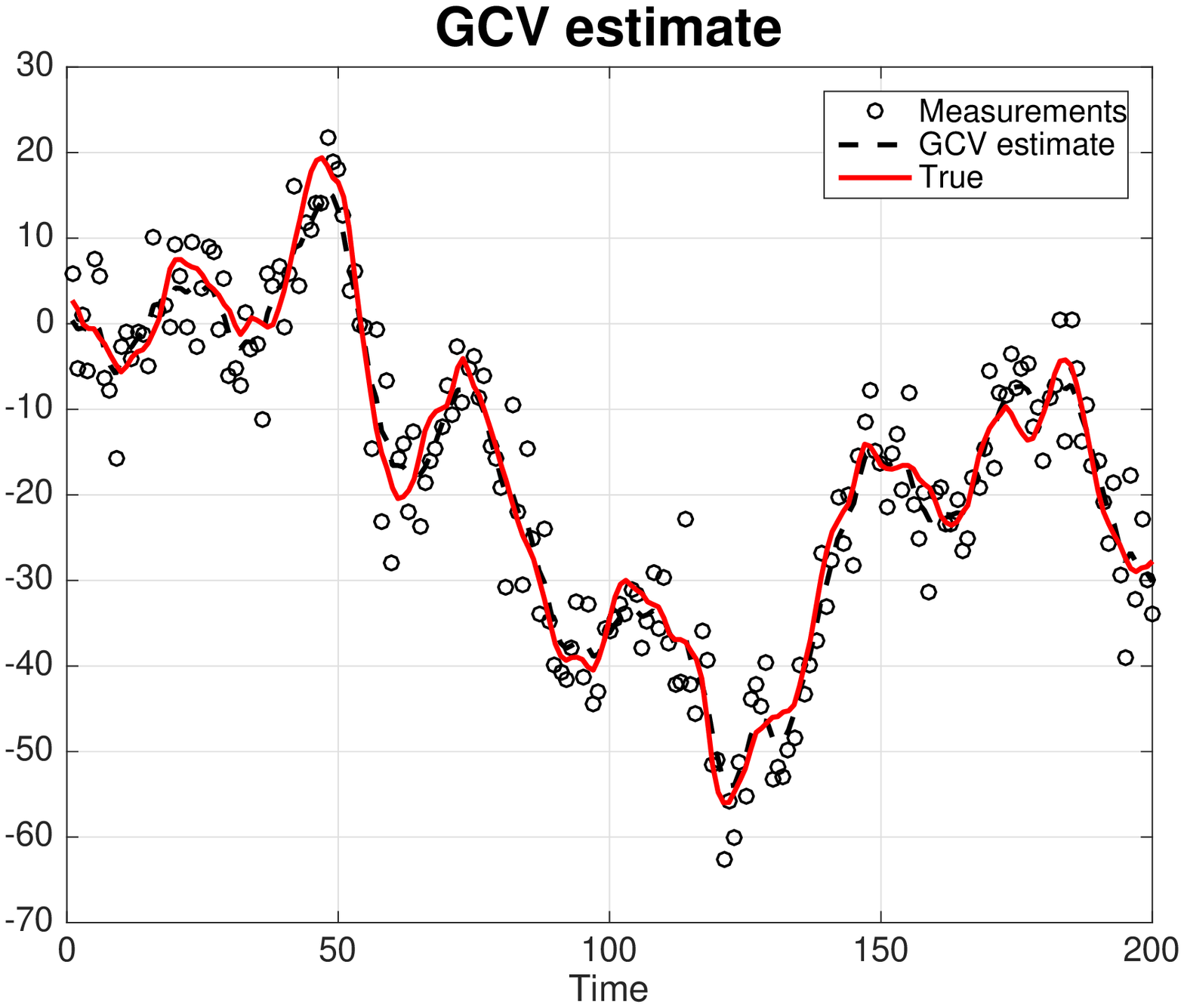}} 
\hspace{.1in}
 { \includegraphics[scale=0.46]{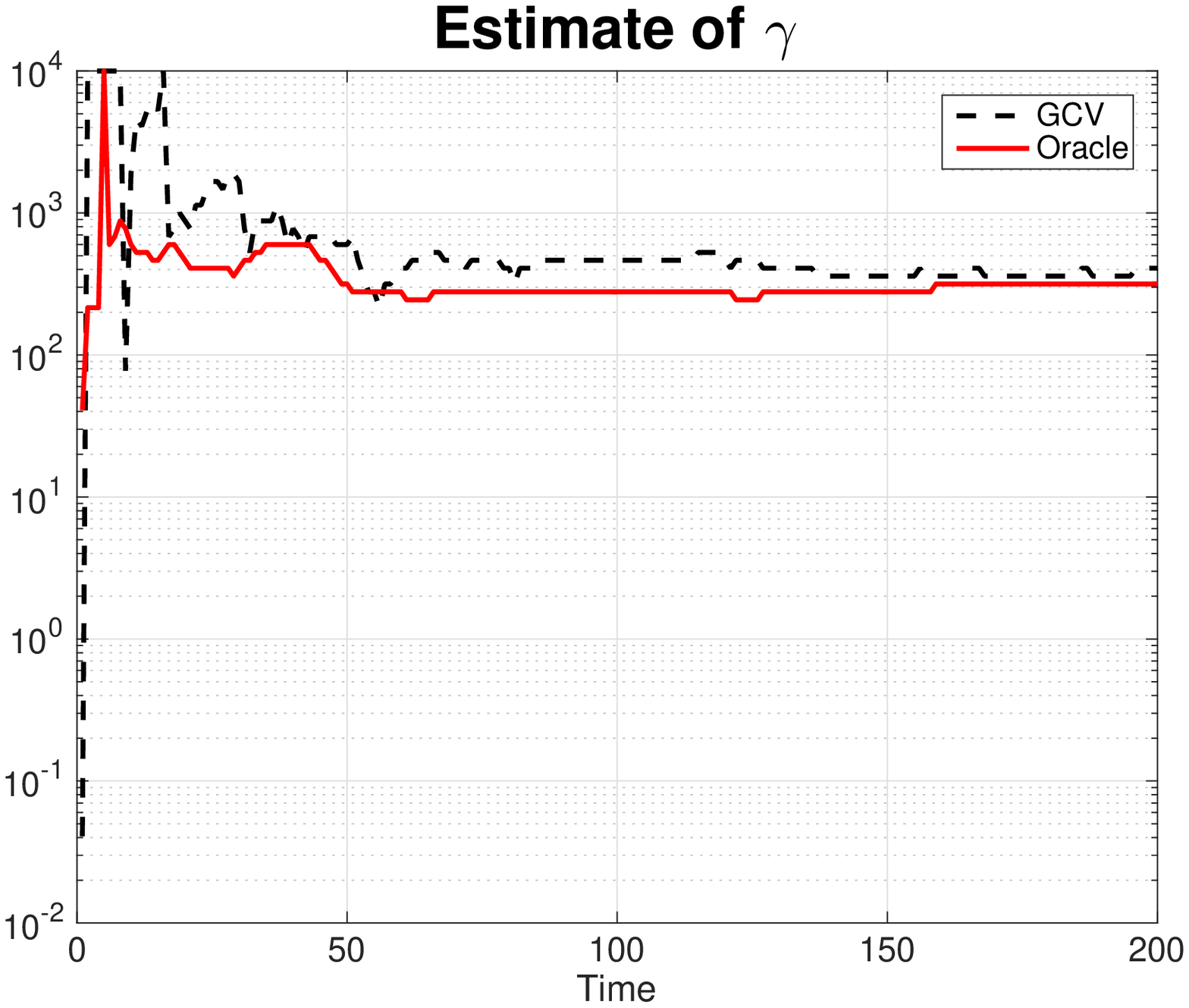}} 
    \end{tabular}
 \caption{{\bf{Model mismatch example - Section \ref{DCmotorsub}}}. {\it{Left:}} noiseless output (solid line), measurements  ($\circ$) and smoothed output obtained by GCV (dashed line). 
 {\it{Right:}} Estimated noise variance $\gamma_t$, as a function of time, obtained by Oracle 
 (solid line) and by GCV (dashed line). } 
    \label{Fig2}
     \end{center}
\end{figure*}

We consider the following discrete-time model (see also \cite[Section 6]{Ohlsson2011}): 
\begin{subequations}\label{DCmotor}
\begin{align} 
x_{k+1} &= \left(\begin{array}{cc} 0.7 & 0 \\0.1 & 1\end{array}\right) x_k
+   \omega_k \\
y_k &= \left(\begin{array}{cc}0 & 1\end{array}\right) x_k + e_k
\end{align}
\end{subequations}
with zero-mean Gaussian noises of covariances 
$$
Q=\left(\begin{array}{c}11.81 \\0.625 \end{array}\right)\left(\begin{array}{c}11.81 \ \ 0.625 \end{array}\right), \quad \gamma=30.
$$
We will use data generated by this model to test the 
capability of the GCV filter to compensate for mismatches between the true system and the model used to track the data
by tuning $\gamma$ in an on-line manner.
As in the previous example
$Z_t$ is the vector containing the 
first $t$ noiseless outputs (which are the second entries of $\{x_k\}_{k=1}^{t}$)
and the performance measure is (\ref{Fit}). 
The following three different estimators $\hat{Z}_t$ are tested:
\begin{itemize}
\item \emph{GCV:} this approach uses a wrong 
transition covariance given by
$$
\tilde{Q}=Q+\left(\begin{array}{cc} 0 & 0 \\0 & 100\end{array}\right),
$$
and then estimates $\gamma$ exploiting the GCV filter over a
grid with 100 values logarithmically spaced on $[10^{-2},10^4]$.
Then, at any $t$ the estimate $\hat{Z}_t$ is computed by a Kalman smoothing filter
which exploits the $\gamma_t$ minimizing $GCV_t$.
\item \emph{Oracle:} the same as GCV except that $\gamma_t$ 
maximizes the fit $\mathcal{F}_t$ in (\ref{Fit}).  
\item \emph{Nominal:} the estimate $\hat{Z}_t$ is returned by a Kalman smoothing
filter defined by the nominal wrong covariance $\tilde{Q}$ and $\gamma=30$. 
Thus, this approach does not try to compensate for model mismatch since it does not tune 
$\gamma$ from data.
\end{itemize}

\begin{figure}[t]
  \begin{center}
   \begin{tabular}{cccc}
\hspace{.1in}
 { \includegraphics[scale=0.46]{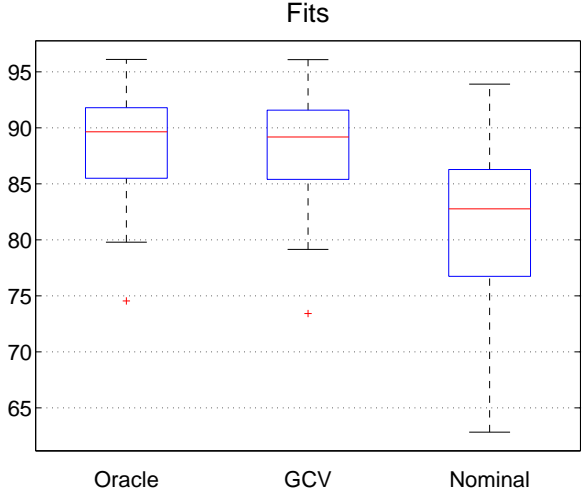}} 
    \end{tabular}
 \caption{{\bf{Model mismatch example - Section \ref{DCmotorsub}}}. Boxplots of the 100 fits 
 $\mathcal{F}_{200}$, as defined in eq. \ref{Fit}, obtained after a Monte Carlo study
 by the estimators Oracle, GCV and Nominal. } 
    \label{Fig3}
     \end{center}
\end{figure}

The left panel of Fig. \ref{Fig2} displays the noiseless output (solid line), the measurements  ($\circ$) 
and the smoothed output obtained by GCV (dashed line) which appears close to truth. In the right panel,
one can also see the trajectory in time of the $\gamma_t$ returned by Oracle
and by GCV. One can appreciate the capability of the GCV filter to compensate the modelling mismatch by tracking 
a regularization parameter leading to a high fit $\mathcal{F}_t$.\\
To further support these findings, we have also performed
a Monte Carlo study of 100 runs. During each run, 200 output measurements are generated using 
(\ref{DCmotor}) and $Z_t$ is reconstructed by GCV, Oracle and Nominal.
From the MATLAB boxplots of the 100 fits (\ref{Fit}) reported in Fig. \ref{Fig3}, the robustness
of GCV emerges clearly. Its performance is in fact very close to that of the oracle-based procedure.

\subsection{On-line regularized linear system identification}\label{OnlineSysid}

Now, we consider  a linear system identification problem where the
aim is to estimate an unknown impulse response from input-output measurements.
Assuming a high order FIR, the model describing
the outputs collected up to instant $t$, and stacked in the (column) vector $Y_t$, 
is 
\begin{equation}\label{FIRmod}
Y_t= \Phi_t g + E_t,
\end{equation}
where $g$ denotes the $m$-dimensional vector 
whose components are the impulse response coefficients, 
the regression matrix $\Phi_t$ is defined by the input samples
and $E_t$ is the measurement noise vector, which we assume white and Gaussian.

To solve this problem, we use the kernel-based approach
originally proposed in \cite{SS2010,SS2011,ChenOL12}.
The impulse response estimate is given by
\begin{equation}\label{SSest}
\argmin_{g \in \R^m} \ \| Y_t - \Phi_t g \|^2 + \gamma g^T P_0^{-1} g.
\end{equation}
It makes use of the regularization matrix $P_0$ induced by the so called
first-order spline kernel, i.e. its $(i,j)$ entry is
$$
[P_0]_{ij} = \alpha^{\max(i,j)}, \quad 0\leq \alpha<1, 
$$
where $\alpha$ is an hyperparameter which regulates the rate of decay to zero of 
the components of $g$. We refer the reader also to \cite{SurveyKBsysid}
for further details on advantages of (\ref{SSest}) over 
classical parametric approaches. 

In real applications, both $\gamma$ and $\alpha$ are unknown.
Since we consider a situation where $g$ has to be estimated on-line, 
we will estimate these two hyperparameters by the GCV filter. 
 {To do that, we first notice that \eqref{SSest} corresponds to the maximum a posteriori (MAP) estimator of $g$ and, under the stated Gaussian assumptions, also to its minimum mean-square estimator (MMSE). The estimate of $g$ can then be computed using the Kalman filter. In fact the state space model is} 
\begin{align} \label{eq:sysid_ss}
x_{k+1} & = x_k \nonumber\\
y_k & = C_k x_k + e_k, \ \ k=1,2,\ldots  \\
x_1 & \sim (0,\,P_0)\nonumber \\ 
e_k & \sim (0,\,\gamma) \nonumber
\end{align}
where the state vector is the stochastic model for $g$ (with $x_k = g$ for any $k$), $y_k$ and $e_k$ are the \Th{k} entries of $Y_t$ and $E_t$, respectively, and $C_k$ is the \Th{k} row of $\Phi_t$. 

We define a grid in the plane $(\gamma,\,\alpha)$ taking the points such that $\alpha$ is in the set $\{0.5,\,0.6,\,\ldots,\,0.9,\,0.95,\,0.99\}$, while $\gamma$ assumes values in a logarithmically spaced grid of 20 point between $10^{-2}$ and $10^3$. In this way, the grid consists of 140 points. We run 140 GCV filters in parallel, each corresponding to one of the points; when a new measure $y_k$ (and $C_k$) is available, we update the GCV score of each pair $(\gamma,\,\alpha)$, selecting the one giving the minimum score.

We test the obtained GCV filter for on-line regularized system identification on a set of 100 Monte Carlo runs. At any run, a random impulse response of length $m=200$ is generated using the same mechanism described in \cite[Section 7.4]{Pillonetto2016}. The generated system is fed with a with noise sequence of unit variance.  {Note that this type of input is persistently exciting and guarantees the observability of the system \eqref{eq:sysid_ss} (see e.g. \cite{And1981}), avoiding the covariance windup phenomenon \cite{stenlund2002avoiding}.}
The standard deviation of the measurement noise is that of the 200 noiseless outputs divided by 10. We assume the system is at rest (the input is equal to zero) prior to the data collection.
The performance of the estimator is evaluated by means of the fit (as a function of time)
\begin{equation}
\mathcal F_t =  100\%\left(1- \frac{\|g^i - \hat g_t^i\|}{\|g^i - 1\!\!\!1 \bar g^i\|}\right),
\end{equation}
where $g^i$ is the impulse response generated at the \Th{i} Monte Carlo run, $\bar g^i$ its mean, and $\hat g_t^i$ its estimate (the impulse response estimate is function of the time instant $t$).

\begin{figure*}
  \begin{center}
   \begin{tabular}{cccc}
\hspace{.1in}
 { \includegraphics[scale=0.46]{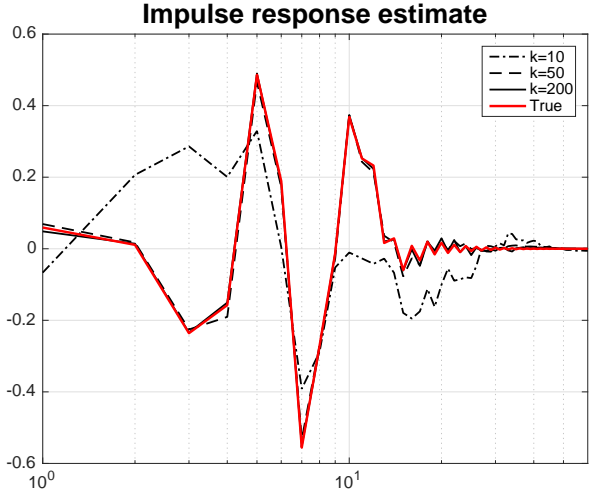}} 
\hspace{.1in}
 { \includegraphics[scale=0.46]{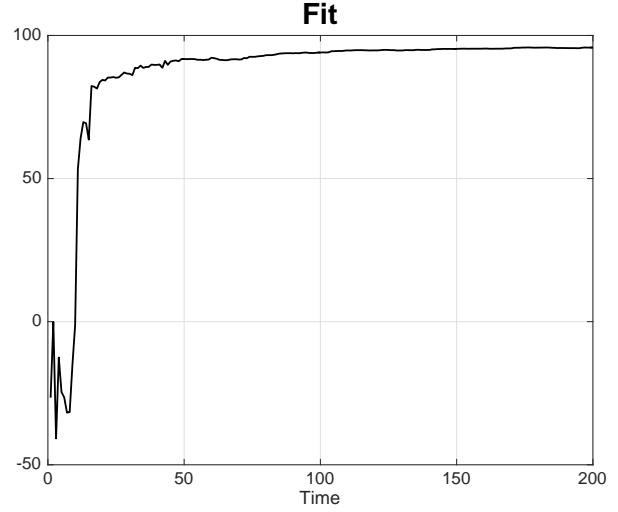}} 
    \end{tabular}
 \caption{{\bf{On-line system identification  - Section \ref{OnlineSysid}}}. {\it{Left:}} true impulse response (solid line) and
 GCV estimates obtained at time instants $k=10,50,200$. 
 {\it{Right:}} Fit obtained by GCV as a function of time. } 
    \label{Fig4}
     \end{center}
\end{figure*}

\begin{figure}
  \begin{center}
   \begin{tabular}{cccc}
\hspace{.1in}
 { \includegraphics[scale=0.46]{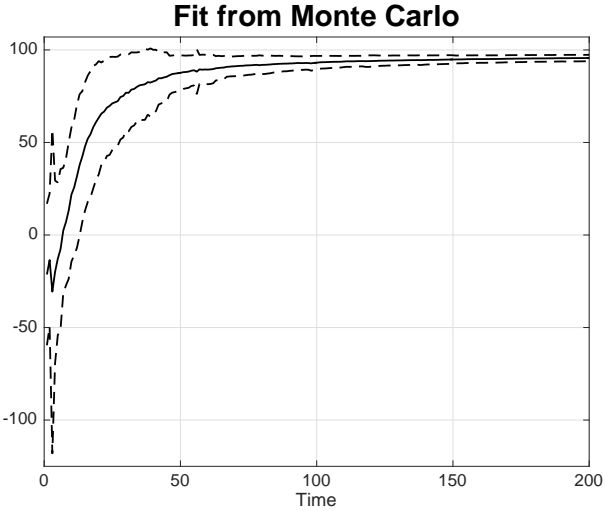}} 
    \end{tabular}
 \caption{{\bf{On-line system identification - Section \ref{OnlineSysid}}}. Average of the GCV fits
 (solid line) $\pm$ one standard deviation (dashed line), as a function of time, obtained after a Monte Carlo study.
 At any of the 100 runs a new impulse response was randomly generated  
as detailed in \cite[Section 7.4]{Pillonetto2016}.} 
    \label{Fig5}
     \end{center}
\end{figure}

An example of one of the Monte Carlo runs is given in Fig. \ref{Fig4}, which shows the evolution in time of the impulse response estimate and its fit. It suffices 50 measurements to the GCV filter to achieve an appreciable fit. The overall results of the Monte Carlo experiment are summarized in Fig. \ref{Fig5}, which depicts the average fit of the impulse responses as a function of time. It can be seen that, after a short transient phase, the fit increases monotonically and achieves a high average value.  

\section{Conclusions}

The novel filter here presented allows to propagate efficiently the 
GCV score over time. Hence, unknown parameters entering a state space model can now be estimated in an
on-line manner resorting to one of the most important techniques used for parameter estimation.
The asymptotic properties of the GCV filter provide also a new very efficient way 
to estimate the regularization parameter  e.g. in smoothing splines.
Applications of the new filter have been illustrated using artificial data
regarding a function estimation and on-line regularized linear system identification. 

 {A Matlab implementation of the GCV filter is available at the web page \texttt{http://www.dei.unipd.it/~giapi/}.}

\section{Appendix}
\subsection{Derivation of the GCV filter}
Without loss of generality, we set the initial system condition to zero, i.e. $\mu=0$.
We also use $X_t,Y_t$ and $E_t$ to denote the column vectors containing
the states, the outputs and the measurements noises up to instant $t$, i.e.
\begin{eqnarray*}
&&  \quad X_t = [x_1^T \ldots x_t^T]^T, \quad Y_t = [y_1 \ldots y_t]^T,
\quad E_t = [e_1 \ldots e_t]^T.
\end{eqnarray*}
Then, it holds that
$$
Y_t = O_t X_t + E_t,
$$
where  {$O_t=\Diag\{C_1,\,\ldots,\,C_t\}$} is the regression matrix built using the measurement matrices $C_k$, $k=1,\ldots,t$.
We also use $W_t$ and $V_t$ to denote the state and output covariance matrix, i.e.   
\begin{eqnarray} \label{St}
W_t &:=&\mbox{Var}(X_t)\\ \label{Vt}
V_t &:=& \mbox{Var}(Y_t) = O_t W_t O_t^T + \gamma I_t, 
\end{eqnarray}
where $I_t$ is the $t \times t$ identity matrix. 
Note that, using the above notation, the 
smoothed estimate of $Y_t$, already encountered in Section \ref{Sec1}, is
\begin{equation}\label{YtA}
\hat{Y}_t = O_tW_tO_t^T  V_t^{-1} Y_t,
\end{equation}
so that the degrees of freedom
at instant $t$ turn out
\begin{equation}\label{DoftA}
\delta_t = \Tr (O_tW_tO_t^T  V_t^{-1}).
\end{equation}
The following simple lemma is useful for our future developments.
\begin{lemma}
One has
\begin{eqnarray}\label{Doft}
\delta_t &=& t - \gamma \frac{\partial \log \det V_t}{\partial \gamma}, \\ \label{Ssrt}
S_t &=&  -\gamma^2  \frac{\partial Y_t  V_t^{-1}  Y_t}{\partial \gamma}. 
\end{eqnarray}
\end{lemma}
{\bf{Proof:}} In view of (\ref{Vt}), we start noticing that
\begin{equation}\label{UsEq}
\gamma V_t^{-1} = I_t -  O_tW_tO_t^T  V_t^{-1}.
\end{equation}
Then, (\ref{Doft}) is obtained from the following 
equalities
\begin{eqnarray} \nonumber
\gamma \frac{\partial \log \det V_t}{\partial \gamma} &=& \gamma \Tr \left (V_t^{-1}\frac{\partial V_t}{\partial \gamma} \right) \\
\nonumber &=& \gamma \Tr (V_t^{-1}) \\
\nonumber &=& \Tr  (I_t- O_t W_t O_t^T  V_t^{-1})  \\
\label{First}  &=& t-\delta_t,
\end{eqnarray}
where the last two passages exploit (\ref{UsEq}) and (\ref{DoftA}), respectively.\\
Eq. \ref{Ssrt} is instead proved as follows
\begin{eqnarray*}
-\gamma^2  \frac{\partial Y_t  V_t^{-1}  Y_t}{\partial \gamma} &=& \gamma^2   Y^T_t  V_t^{-2}  Y_t \\
&=& Y_t^T (I_t -  O_tW_tO_t^T  V_t^{-1} )^T(I_t -  O_tW_tO_t^T  V_t^{-1} ) Y_t \\
&=& \| Y_t - \hat{Y}_t \|^2 = S_t
\end{eqnarray*}
where  the second and third equality exploit (\ref{UsEq}) and (\ref{YtA}), respectively.\\
\begin{flushright}
$\blacksquare$
\end{flushright}

The dynamics of the matrix $P_k$ in the GCV filter are regulated
by the discrete-time algebraic Riccati equation (DARE), which can be also rewritten as
$$
P_{k+1} = A_{k} P_k A_{k}^T + Q_{k} - A_{k}P_kC_{k}^T(C_{k} P_k C_{k}^T+\gamma)^{-1}C_{k}P_kA_{k}^T.
$$
It is now easy to see that the matrix $\Sigma_k$ entering the GCV filter is the partial derivative of
$P_k$ w.r.t. $\gamma$. In fact, 
differentiating the DRE, and adopting the notation $\Sigma_k := \frac{\partial P_k}{\partial \gamma}$, one has 
\begin{eqnarray*}
\Sigma_{k+1} &=& A_{k}\Sigma_kA_{k}^T - A_{k}\Sigma_kC_{k}^T(C_{k} P_k C_{k}^T+\gamma)^{-1}C_{k}P_kA_{k}^T \\
&-& A_{k}P_kC_{k}^T (C_{k} P_k C_{k}^T+\gamma)^{-1} C_{k}\Sigma_kA_{k}^T \\
&+& A_{k}P_kC_{k}^T(C_{k} P_k C_{k}^T+\gamma)^{-2}C_{k}P_kA_{k}^T(C_{k} \Sigma_k C_{k}^T+1).
\end{eqnarray*}
Exploiting the definition of $K_k$ and rearraging the terms, the recursive
formula (\ref{DRE2}) is obtained. 

Now, consider the dynamics of the predicted state 
$$
\hat{x}_{k+1} = A_{k} \hat{x}_{k} + A_{k}P_{k}C_{k}^T  (C_{k} P_{k} C_{k}^T +\gamma)^{-1} (y_{k} - C_{k} \hat{x}_{k} ).
$$
We now show that $\hat{\zeta}_{k}$ is the partial derivative of
$\hat{x}_k$ w.r.t. $\gamma$. In fact,   
differentating the above equation using the correspondence $\hat{\zeta}_k := \frac{\partial \hat{x}_{k}}{\partial \gamma}$,
one obtains 
\begin{eqnarray*}
\hat{\zeta}_{k+1} &=& A_{k} \hat{\zeta}_k + A_{k} \Sigma_k C_{k}^T (C_{k} P_{k} C_{k}^T +\gamma)^{-1}  (y_{k} - C_{k} \hat{x}_{k} )\\
&-& A_{k} \Sigma_k C_{k}^T (C_{k}\Sigma_k C_{k}^T+1)(C_{k} P_{k} C_{k}^T +\gamma)^{-2} (y_{k} - C_{k} \hat{x}_{k} )\\
&-& A_{k}P_k C_{k}^T (C_{k} P_{k} C_{k}^T +\gamma)^{-1} C_{k} \hat{\zeta}_k.
\end{eqnarray*}
This, combined with the definition of $K_k$, leads to the recursive
formula (\ref{zpred}). 

Now, exploiting well known properties of the innovations sequence
$\{y_k-C_{k}\hat{x}_k\}_{k=1}^t$, whose variances are $\{C_{k} P_k C_{k}^T + \gamma\}_{k=1}^t$, and 
recalling that $\Sigma_k := \frac{\partial P_k}{\partial \gamma}$, we have
\begin{eqnarray*}
\frac{\partial  \log \det V_t}{\partial \gamma} &=& \sum_{k=1}^t \frac{\partial \log ( C_{k} P_k C_{k}^T + \gamma )}{\partial \gamma}  \\
&=& \sum_{k=1}^t \frac{C_{k} \Sigma_k C_{k}^T+1}{C_{k} P_k C_{k}^T+\gamma}.
\end{eqnarray*}
Then, the recursive formula (\ref{DofRec})
for the degrees of freedom $\delta_k$  is obtained combining the  
above equation and (\ref{Doft}).\\

Still using properties of the innovations sequence, and recalling that
$\hat{\zeta}_k := \frac{\partial \hat{x}_{k}}{\partial \gamma}$, one also has
 \begin{eqnarray*}
 -\frac{\partial Y_t  V_t^{-1}  Y_t}{\partial \gamma} &=& -\sum_{k=1}^t \ \frac{\partial (y_k-C_{k}\hat{x}_k)^2  ( C_{k} P_k C_{k}^T + \gamma )^{-1}}{\partial \gamma} \\
 &=& \sum_{k=1}^t   \frac{C_{k} \Sigma_{k} C_{k}^T + 1}{(C_{k} P_{k} C_{k}^T +\gamma)^{2}}(y_{k} - C_{k} \hat{x}_{k})^2 \\
 &+&  2C \hat{\zeta}_{k} \frac{y_{k} - C_{k} \hat{x}_{k}}{C_{k} P_{k} C_{k}^T +\gamma}.
\end{eqnarray*}
This equation, in combination with (\ref{Ssrt}), proves the correctness of the update rule
(\ref{SsrRec}) for
the sum of squared residuals $S_k$ and completes the derivation.

 {
\subsection{Proof of Proposition \ref{GCVregime}}
If the system (\ref{StateMod}) is stabilizable and detectable, then standard properties of the algebraic Riccati equation \eqref{ARE} ensure that $\bar{P}$ is symmetric and positive semidefinite and that the Kalman filter, corresponding to \eqref{Kgain}, \eqref{xpred} and \eqref{DRE}, is asymptotically stable (see \cite[p. 77]{Anderson:1979}). Because the Kalman filter is asymptotically stable, the matrix $(A-\bar{K}C)$ has all the eigenvalues inside the unit circle, ensuring that \eqref{ARE2} admits a unique positive semidefinite solution \cite[p. 67]{Anderson:1979}. The filter state transition matrix which regulates the dynamics of
$\hat{x}_k$ and $\hat{\zeta}_k$  is
$$
\left(\begin{matrix}
A - K_k C     &   0 \\
-G_k C     &     A - K_k C
\end{matrix}\right)
$$
and so it also has all eigenvalues inside the unit circle at least for sufficiently large $k$. 
}

\bibliographystyle{plain}
\bibliography{biblio}
\end{document}